\documentclass[11pt]{article}

\usepackage{acl}

\usepackage{times}
\usepackage{latexsym}
\usepackage{amsmath}
\usepackage{booktabs}
\usepackage{multirow}
\usepackage{array}
\newcommand{\midfont}{\fontsize{8.3}{10.2}\selectfont}
\usepackage[T1]{fontenc}
\usepackage[utf8]{inputenc}

\usepackage{microtype}

\usepackage{inconsolata}

\usepackage{graphicx}

\title{Decoupling Patient Personas and Clinical Workflows: A Dual-Agent Approach to Realistic Medical Dialogue Generation}

\author{
  \textbf{Qiyuan Li\textsuperscript{1}},
  \textbf{Xiaomei Guo\textsuperscript{2}},
  \textbf{Chao Gao\textsuperscript{3}},
  \textbf{Haijiang Liu\textsuperscript{1}}, \\
  \textbf{Deyu Chen\textsuperscript{4}},
  \textbf{Caicai Guo\textsuperscript{1}},
  \textbf{Jinguang Gu\textsuperscript{1,*}} \\
  \\
  \textsuperscript{1}School of Computer Science and Technology,
  Wuhan University of Science and Technology, China \\
  \textsuperscript{2}Department of Cardiology, Tongji Hospital, Huazhong University of Science and Technology, China \\
  \textsuperscript{3}Department of Computer Science,
  Vrije Universiteit Amsterdam, Netherlands \\
  \textsuperscript{4}School of Computer Science and Technology,
  Huazhong University of Science and Technology, China \\
  {\small
    \textbf{First author:} \texttt{vickyuan@wust.edu.cn} 
    \textbf{*Corresponding author:} \texttt{simon@wust.edu.cn}
  }
}

\begin{document}
\maketitle
\begin{abstract}
Pre-consultation medical dialogues are essential for collecting comprehensive patient information, but are costly to obtain at scale due to privacy and ethical constraints. Existing LLM-based synthesis methods often rely on prompt engineering or rigid rules, limiting their ability to balance clinical relevance with realistic patient diversity.
We propose \textbf{CoRPS-Dial}, a controllable dual-agent framework for pre-consultation medical dialogue synthesis that integrates patient personas with guideline-aligned clinical workflow rules. CoRPS-Dial generates dialogues through interactions between a persona-driven patient agent and a process-guided doctor agent, followed by post-hoc quality control for role consistency, workflow completeness, and textual consistency.
Experiments on coronary heart disease demonstrate that CoRPS-Dial improves alignment between synthetic and real dialogues in information coverage, clinical compliance, and dialogue naturalness. Downstream fine-tuning experiments further confirm the utility of the generated synthetic data for medical dialogue tasks. These results highlight the effectiveness of CoRPS-Dial for clinically grounded and controllable medical dialogue synthesis.
% \footnote{https://anonymous.4open.science/r/CoRPS-Dial-d2c6/}
\end{abstract}

\section{Introduction}
\label{intro}
Pre-consultation medical dialogues are crucial for gathering comprehensive patient information for effective triage and diagnostic decision-making \citep{Yu2025}. High-quality multi-turn dialogues are also valuable for training medical dialogue systems to emulate diagnostic reasoning. However, acquiring large-scale real-world doctor–patient dialogues is challenging due to privacy regulations, ethical constraints, and expensive expert annotation, particularly for specialized diseases with complex clinical workflows and strict information requirements.

\begin{figure}[t]
  \includegraphics[width=\columnwidth]{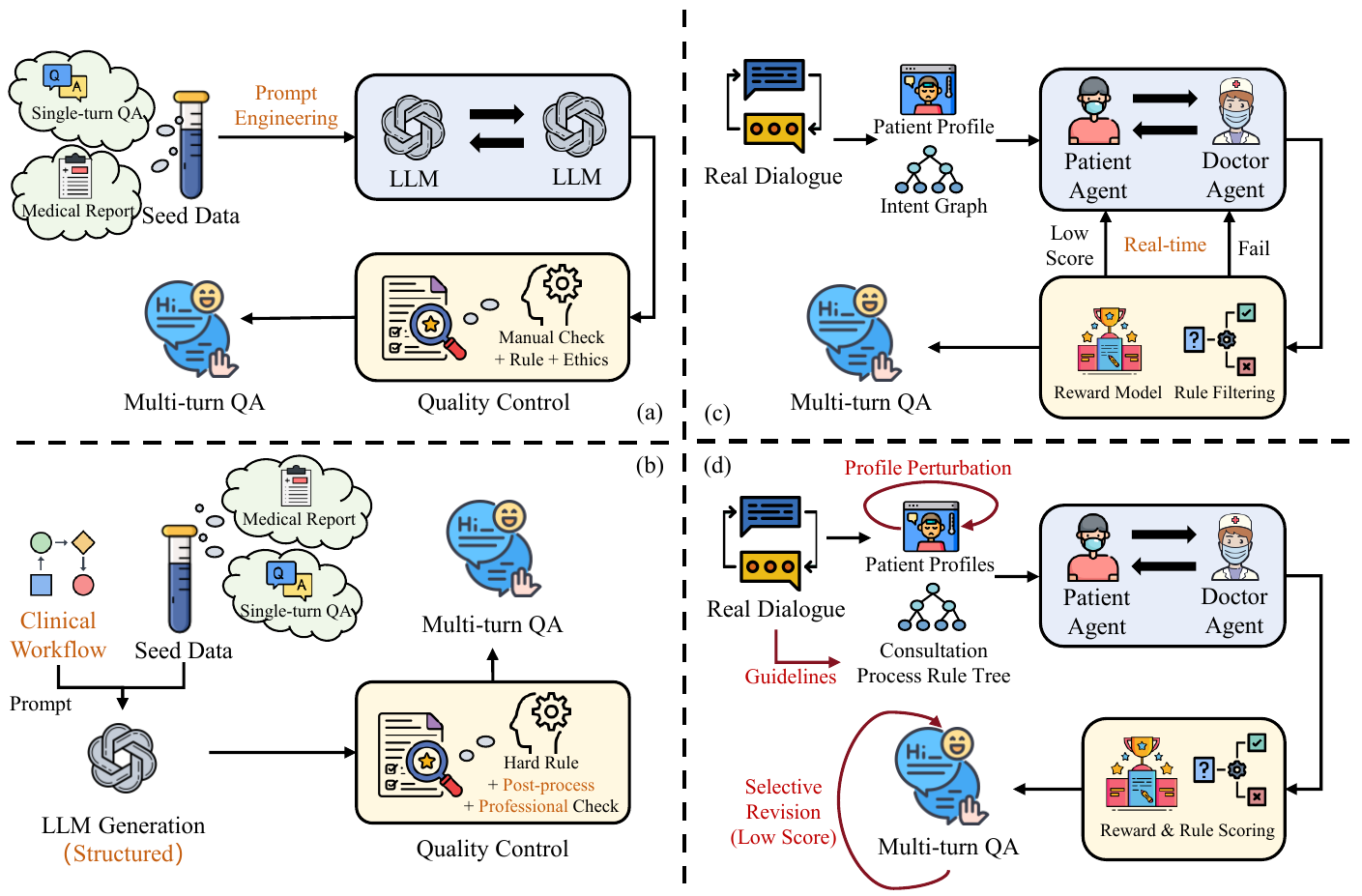}
  \caption{Overview of dialogue generation approaches. (a) Prompt-driven generation with limited clinical reliability. (b) Clinical workflow-driven generation with costly expert verification. (c) Intent-driven multi-agent simulation with real-time rule validation and iterative optimization requiring repeated LLM calls. (d) CoRPS-Dial, a patient persona and clinical process dual-driven framework with post-hoc validation and selective refinement.}
  \label{fig:fig1}
\end{figure}

To alleviate dialogue scarcity, recent efforts have leveraged large language models (LLMs) to generate or assist in generating medical dialogues \citep{smile, PsyDT, SoulChat, CPsyCoun}, typically through prompt engineering or template-based designs (Figure~\ref{fig:fig1}(a)). While these approaches demonstrate strong multi-turn generation ability, their clinical reliability is limited by implicit parametric knowledge and hallucinated content. To improve controllability, prior work incorporates explicit clinical rules \citep{RuleAlign} (Figure~\ref{fig:fig1}(b)) or hybrid rule–reward mechanisms \citep{Ge2025} (Figure~\ref{fig:fig1}(c)). However, existing methods still face a trade-off between scalability and reliability: fine-grained control often depends on expert-designed rules and manual verification, while iterative multi-turn control increases pipeline complexity and may reduce dialogue consistency. These limitations are particularly critical in pre-consultation settings, where patient presentations exhibit substantial heterogeneity in symptoms, emotions, and communication styles, while pre-consultation interviews must follow structured, guideline-informed workflows. Existing approaches couple these objectives within a single generation process, limiting independent control.

To address these challenges, we propose a controllable dual-agent framework, CoRPS-Dial, that explicitly decouples patient heterogeneity from guideline-aligned clinical process modeling (Figure~\ref{fig:fig1}(d)). Specifically, patient personas are constructed from a seed set of real, de-identified cardiology consultations transcribed from clinical audio recordings, capturing demographic attributes, health conditions, emotions, and communication styles. In parallel, guideline-aligned clinical process rules explicitly encode workflow structure and clinical priorities. Dialogue generation is realized through interactions between a persona-driven patient agent and a process-guided doctor agent, followed by a post-hoc quality control pipeline that ensures medical consistency and workflow completeness via rule-based checks, reward scoring, and selective refinement. Experiments on coronary heart disease (CHD) demonstrate the effectiveness of CoRPS-Dial for controllable and clinically grounded medical dialogue synthesis. Downstream fine-tuning further validates the utility of the generated synthetic data for medical dialogue tasks.

Our contributions are summarized as follows:
\begin{itemize}
    \item We introduce CoRPS-Dial, a dual-agent framework for controllable pre-consultation medical dialogue synthesis under dialogue scarcity, which explicitly decouples patient heterogeneity modeling from guideline-informed clinical process control.
    \item We design a structured generation and post-hoc quality control pipeline that ensures medical consistency, workflow completeness, and realistic patient heterogeneity.
    \item A CHD case study shows that CoRPS-Dial generates synthetic dialogues that are better aligned with real-world pre-consultation dialogues in terms of structure, clinical workflow adherence, and coherence. Downstream fine-tuning further demonstrates the utility of the generated synthetic data for medical dialogue tasks.
\end{itemize}

\section{Related Work}
\label{related-work}
Current medical dialogues can be broadly categorized into two main types: those derived from real doctor–patient interactions, and those generated through human annotation or model-based synthesis \citep{Li2025}.

\subsection{Real-World Medical Dialogues}
Real-World Medical Dialogues are typically sourced from clinical practices or online medical consultation platforms, capturing authentic doctor–patient interactions and natural language expression habits. Representative dialogues include WebMedQA \citep{WebMedQA}, cMedQA2 \citep{cMedQA2}, iCliniq-10k \citep{chatdoctor}, HealthCareMagic\-100k \citep{chatdoctor}, MedDialog \citep{MedDialog}, Huatuo-26M \citep{Huatuo-26M}, Chinese medical dialogues\footnote{\url{https://github.com/Toyhom/Chinese-medical-dialogue-data}} covering six specialties, and CMtMedQA \citep{Zhongjing}, which spans 14 medical departments and multiple medical question-answering (Q\&A) scenarios. These dialogues offer clear advantages in terms of coverage across medical domains and scales. However, most focus on general or multi-disciplinary scenarios and mainly consist of single-turn Q\&A pairs or weakly structured interactions, making them insufficient for disease-specific pre-consultation scenarios requiring specialized medical knowledge and strict clinical workflows. In addition, privacy and ethical constraints limit the availability, scalability and reusability of authentic doctor–patient dialogues.

\begin{figure*}[t]
  \centering
  \includegraphics[width=\linewidth]{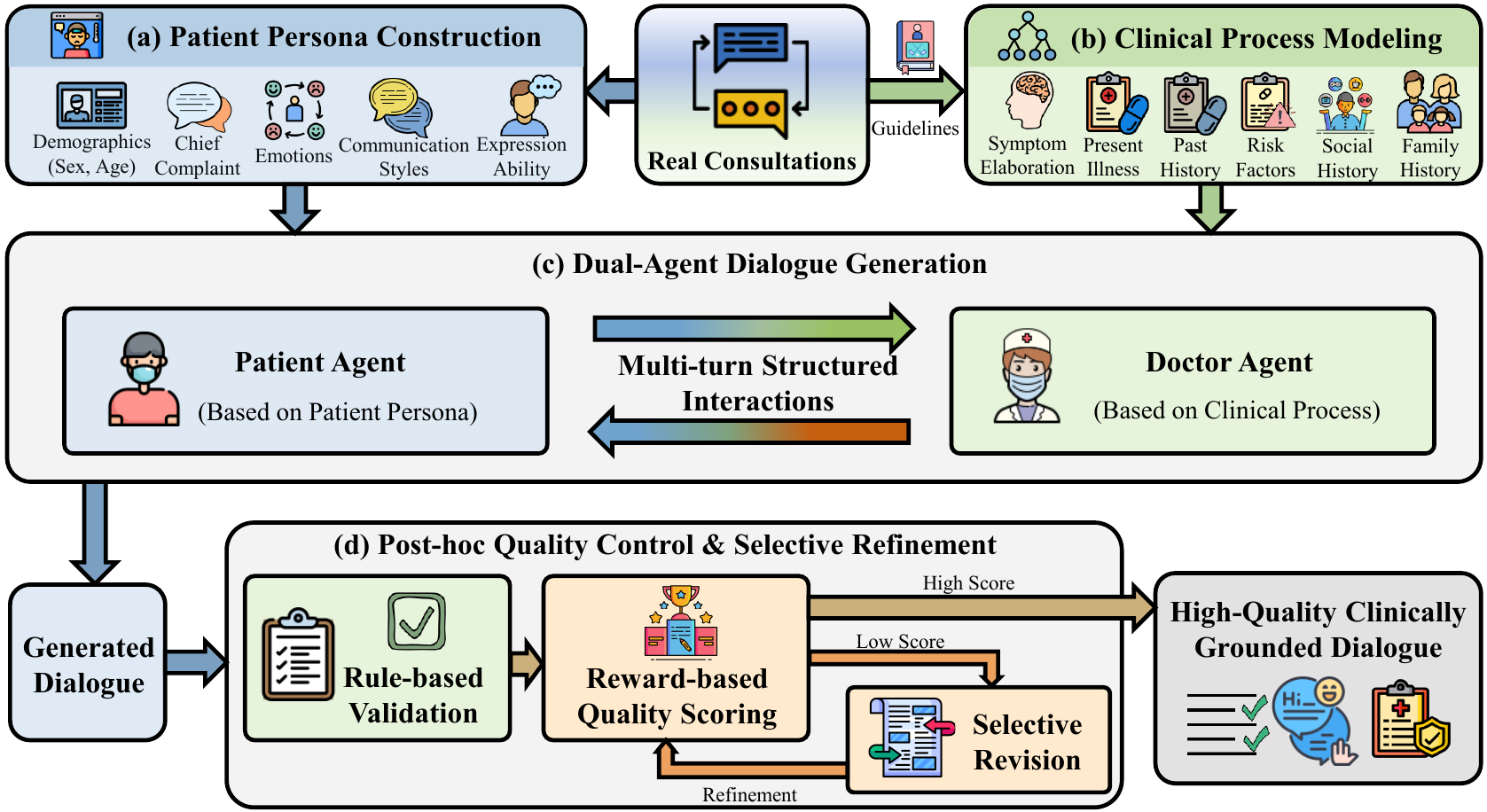}
  \caption {CoRPS-Dial Framework. \textbf{(a) Patient Persona Construction}, deriving diverse profiles from real consultations capturing demographics, chief complaints, emotions, communication styles, and expression abilities; \textbf{(b) Clinical Process Modeling}, abstracting real consultations with guidelines into symptom elaboration, present illness history, past history, risk factors, social history and family history; \textbf{(c) Dual-Agent Dialogue Generation}, enabling multi-turn structured interactions between a persona-based patient agent and a process-based doctor agent; \textbf{(d) Post-hoc Quality Control \& Selective Refinement}, applying rule-based validation, reward-based scoring, and targeted revisions for high-quality, grounded clinical dialogues.}
  \label{fig:fig2}
\end{figure*}

\subsection{Synthetic Medical Dialogues}
Synthetic Medical Dialogues have recently been explored to alleviate the scarcity of real-world dialogues, with LLMs used to generate or assist in generating them. LLM-generated dialogue offers stronger controllability, lower privacy risk, and greater scalability through prompt engineering or template-based designs. Prior studies such as ChatMed-Consult \citep{ChatMed}, GenMedGPT-5k \citep{chatdoctor}, and RJUA-QA \citep{RJUA-QA} generate medical Q\&A pairs or simulated consultations, while more recent works, including SmileChat \citep{smile}, PsyDTCorpus \citep{PsyDT}, SoulChatCorpus \citep{SoulChat}, and CPsyCounD \citep{CPsyCoun}, extend synthesis to multi-turn interactions. However, these approaches remain limited by hallucinations and weak clinical grounding, which can reduce clinical reliability and professional accuracy.

To improve clinical plausibility, recent studies incorporate clinical rules or hybrid rule-and-reward quality control mechanisms. For example, Wang et al. \citep{RuleAlign} transform single-turn Q\&A into workflow-aligned multi-turn dialogue skeletons with rule constraints and expert review, while Ge et al. \citep{Ge2025} combine intent planning, dual-agent simulation, and rule–reward mechanisms for step-wise medical dialogue generation. However, these methods often rely heavily on expert involvement or lack explicit alignment with standardized clinical guidelines, limiting scalability and robustness in disease-specific pre-consultation scenarios with complex workflows. Moreover, enforcing medical constraints during online generation can reduce global dialogue consistency.

To address these challenges, we propose a dual-agent framework that integrates patient personas with guideline-aligned workflow rules to achieve controllable and clinically reliable pre-consultation dialogue synthesis.

\section{Methodology}
\label{method}
In this section, we introduce CoRPS-Dial, a dual-agent framework for controllable generation of clinically grounded multi-turn dialogues, illustrated in Figure~\ref{fig:fig2}.

\subsection{Problem Formulation}
\label{sec:problem-formulation}
% Our goal is to automatically generate multi-turn doctor–patient dialogues that (1) follow valid clinical workflows, (2) collect clinically critical information with appropriate priorities, and (3) reflect realistic patient heterogeneity.

Formally, let $\mathcal{G}$ denote clinical guidelines, defining inquiry items and recommended examinations. The pre-consultation workflow is modeled as a rule tree $\mathcal{T}$, where nodes correspond to medical inquiries, and edges capture dependencies and priority. Patient profiles are represented by $\mathcal{P}$, capturing demographic attributes, health conditions, emotions, and communication styles. A patient instance $p \in \mathcal{P}$ specifies the latent characteristics that govern patient responses during the dialogue.

A pre-consultation dialogue $\mathcal{D}$ is defined as a sequence of turns:

\begin{equation}
  \label{eq:eq1}
  \mathcal{D} = \{(u_t^{d}, u_t^{p})\}_{t=1}^{T}
\end{equation}
where $u_t^{d}$ and $u_t^{p}$ denote the doctor and patient utterances at turn $t$, respectively. The Doctor Agent selects an inquiry based on the rule tree $\mathcal{T}$, and the Patient Agent generates a response conditioned on the inquiry and sampled profile $p$.

Our objective is to generate $\mathcal{D}$ that satisfies: (1) procedural completeness with respect to $\mathcal{T}$, (2) clinical consistency with $\mathcal{G}$, and (3) naturalness while explicitly controlling patient variability via $p$. CoRPS-Dial decouples the disease-specific workflow from disease-agnostic patient profiles, enabling flexible adaptation across medical domains.

\subsection{Patient Persona Construction}
\label{sec:persona-construction}
Real-world patients exhibit substantial heterogeneity even within the same disease, varying in demographics, symptoms, emotions, communication styles, and expressive ability. CoRPS-Dial introduces a patient profiling module that models these attributes as structured latent variables for controllable generation.

We construct a seed dialogue set in collaboration with a tertiary hospital cardiology department using real outpatient pre-consultation recordings. Audio recordings are transcribed with automatic speech recognition and structured by DeepSeek-R1-14B \citep{deepseek-r1} for role identification and turn segmentation. Both models are deployed locally to ensure privacy compliance. Six trained annotators manually verify the transcripts against the original recordings, producing a high-quality de-identified seed dialogue set grounded in real clinical practice. To preserve clinically meaningful patient heterogeneity, we retain informative signals while filtering surface-level noise and irregularities (Appendix~\ref{nosie}).

We define a patient profile $p$ as a tuple:
\begin{equation}
  \label{eq:eq2}
    p = \{p^{\text{demo}}, p^{\text{sym}}, p^{\text{emo}}, p^{\text{comm}}, p^{\text{exp}}\}
\end{equation}
where $p^{\text{demo}}$ captures demographic attributes (e.g., age, gender). $p^{\text{sym}}$ encodes symptoms. $p^{\text{emo}}$ reflects emotional states (e.g., happiness, sadness). $p^{\text{comm}}$ describes communication traits, including verbosity and clarity of expression. And $p^{\text{exp}}$ denotes the primary source of the narrative — self (patient), family, or mixed. These dimensions are largely disease-agnostic and can be instantiated using limited de-identified dialogue statistics and clinical priors.

Due to limited annotated patient attributes in real-world dialogues, we extract de-identified profiles and expand $\mathcal{P}$ via controlled perturbations. Age is varied within $\pm 5$ years \citep{Zhao2024, Knuuti2020}, while other attributes are perturbed by random sampling from the full attribute sets ($\mathcal{P}^{\text{sym}}, \mathcal{P}^{\text{emo}}, \mathcal{P}^{\text{comm}}, \mathcal{P}^{\text{exp}}$) to maintain clinical plausibility and diversity.

A patient profile $p'$ is sampled from this augmented profile space $\mathcal{P'}$, and it conditions the Patient Agent’s responses at every dialogue turn:
\begin{equation}
  u_t^{p'} \sim \mathcal{P'} (u \mid u_t^{d}, h_t, p') \label{eq:patient}
\end{equation}
where $h_t$ is the dialogue history up to turn $t$, and $u_t^{d}$ denotes the doctor's utterance at turn $t$.

\subsection{Clinical Process Modeling}
\label{sec:clinical-process-modeling}
To ensure that generated dialogues are both clinically reliable and conversational, CoRPS-Dial models the pre-consultation inquiry process using a guideline-grounded inquiry tree $\mathcal{T}$. We parse clinical diagnostic guidelines to define a schema $\mathcal{I}$:

\begin{multline}
\label{eq:clini-process}
\mathcal{I} = \{\mathcal{I}^{\text{Gen}}, \mathcal{I}^{\text{Chi}}, \mathcal{I}^{\text{Pre}}, 
\mathcal{I}^{\text{Pas}}, \mathcal{I}^{\text{Rev}}, \\
\mathcal{I}^{\text{Per}}, \mathcal{I}^{\text{Mar}}, \mathcal{I}^{\text{MCH}}, \mathcal{I}^{\text{Fam}}\}
\end{multline}
where $\mathcal{I}^{\text{Gen}}$ represents general information (e.g., age, gender). $\mathcal{I}^{\text{Chi}}$ represents the chief complaint. $\mathcal{I}^{\text{Pre}}$ represents present illness. $\mathcal{I}^{\text{Pas}}$ represents past history. $\mathcal{I}^{\text{Rev}}$ represents review of systems. $\mathcal{I}^{\text{Per}}$ represents personal history (or social history). $\mathcal{I}^{\text{Mar}}$ represents marital history. $\mathcal{I}^{\text{MCH}}$ represents menstrual history and childbearing history (primarily for female patients). $\mathcal{I}^{\text{Fam}}$ represents family history.

To complement guideline-defined workflows with disease-specific nuances and follow-up patterns, we mine frequent inquiry transitions and conditional triggers from real-world consultations $D_{\text{real}}$ (size $N$, covering $M$ cases of the target disease). The resulting tree $\mathcal{T}$ preserves mandatory guideline items and safety constraints while augmenting them with empirically validated optional branches, improving alignment with real clinical practice. During dialogue generation, the doctor agent traverses $\mathcal{T}$ to select the next inquiry, prioritizing mandatory nodes while allowing flexible LLM-based realization, thereby decoupling workflow control from language generation.

\subsection{Dual-Agent Dialogue Generation}
\label{sec:dual-agent}
At each turn $t$, the doctor agent observes the dialogue state $h_t$ (including history and node completion status in $\mathcal{T}$) and selects an inquiry action $a_t$ corresponding to an active node. The policy prioritizes mandatory, high-risk, and prerequisite nodes to ensure workflow adherence. The selected action is then realized as a natural-language question via a language model conditioned on the node semantics.

The patient agent generates a response conditioned on the doctor's inquiry $u_t^{d}$, the dialogue history $h_t$, and the fixed patient profile $p'$ (Section~\ref{sec:persona-construction}), following Equation~\ref{eq:patient}, ensuring consistent persona behavior such as symptoms, emotion, and verbosity.

The dialogue terminates when all mandatory nodes in $\mathcal{T}$ are covered or guideline-defined stopping criteria are met. By decoupling decision-level control from surface realization, the framework improves medical reliability while maintaining naturalness and diversity.

\subsection{Post-hoc Quality Control \& Selective Refinement}

\subsubsection{Hard \& Soft Quality Control}
We perform a hybrid hard--soft evaluation to optimize low-quality outputs efficiently.

First, a deterministic rule-based validator enforces strict compliance on four critical dimensions:

\begin{itemize}
    \item \textbf{Role Consistency}: Whether the questions and answers align with the roles of the patient, doctor, or family member.
    \item \textbf{Process Consistency}: Whether the process aligns with mandatory nodes in $\mathcal{T}$, follows the correct order without violating red flags or dependencies.
    \item \textbf{Contextual Coherence}: Whether the dialogue history contains contradictions in terms of symptom facts, timelines, or medical logic.
    \item \textbf{Basic Text Quality}: Whether there are any formatting errors, incorrect medical terms, unsafe content, or violations of guidelines $\mathcal{G}$.
\end{itemize}
Violations trigger immediate flagging for revision.

Second, we use the GLM-4-9B \citep{glm4} to score the entire dialogue across the same dimensions on a 1–10 Likert scale (details in Appendix~\ref{glm4-score}). If the average reward $R(\mathcal{D})$ is below a threshold $\theta$ (average reward score of real dialogues under the same dimensions and GLM-4-9B scoring), the dialogue is selected for refinement.

This hard-first, soft-second pipeline eliminates critical errors while leveraging GLM-4-9B’s strong discriminative ability and effectiveness in reward modeling for nuanced assessment \citep{LongReward}, achieving high reliability with modest compute overhead.

\subsubsection{Selective Utterance Revision}
For flagged utterances, we perform targeted revision instead of full regeneration. An LLM is prompted with the original utterance, dialogue context $h_t$, patient profile $p'$, and specific error feedback from the assessment stage (e.g., ``resolve symptom inconsistency between turns 3 and 5''). The revised utterance replaces the original only if its reward score improves. This iterative, selective strategy preserves dialogue structure, minimizes compute cost, and yields clinically grounded, high-quality synthetic dialogues.

\section{Experiments}
\label{exper:set}

\begin{table*}
\centering
% \footnotesize
\midfont
% \footnotesize
\renewcommand{\arraystretch}{1.1}
\begin{tabular}{lcccccc}
\hline
Metric & CoRPS-Dial & Raw Doctor & Free Patient & Single Agent & Minimal Agents & No QC/SR \\
\hline
\multicolumn{7}{c}{\textbf{Structure}} \\
\hline
Turns/Dialog & 44.00 ± 0.52 & 18.37 ± 2.41 & 41.26 ± 1.13 & 43.12 ± 1.39 & 27.84 ± 6.92 & 44.00 ± 0.00 \\
Sent Len & 52.99 ± 64.00 & 21.43 ± 18.92 & 47.85 ± 51.76 & 21.11 ± 8.21 & 38.14 ± 42.07 & 66.27 ± 158.96 \\
Unique Chars & 1106 & 742 & 983 & 579 & 1287 & 1209 \\
Tokens/Turn & 36.51 & 15.82 & 33.94 & 13.77 & 29.17 & 44.41 \\
\hline
\multicolumn{7}{c}{\textbf{Semantic}} \\
\hline
Local Coh & 0.5062 ± 0.0123 & 0.3728 ± 0.0216 & 0.4415 ± 0.0189 & 0.5603 ± 0.0217 & 0.3186 ± 0.0473 & 0.5105 ± 0.0197 \\
\hline
\multicolumn{7}{c}{\textbf{Diversity}} \\
\hline
Dist-1 & 0.0490 & 0.0367 & 0.0412 & 0.0317 & 0.0579 & 0.0439 \\
Dist-2 & 0.2371 & 0.1820 & 0.2012 & 0.1326 & 0.2914 & 0.2323 \\
Entropy & 8.2163 & 7.6841 & 7.9327 & 7.1251 & 8.7415 & 8.3086 \\
Total Words & 44,597 & 8,731 & 39,842 & 28,395 & 23,966 & 55,163 \\
\hline
\end{tabular}
\caption{Dialogues generation quality comparison of structure, semantic, and diversity. ``Sent Len'' means sentence length. ``Local Coh'' means local coherence. ``xxx ± xxx'' denotes mean ± standard deviation (std).}
\label{tab:abl}
\end{table*}

\begin{table*}[tb]
\centering
\midfont
\renewcommand{\arraystretch}{1.1}
\begin{tabular}{llcccc}
\hline
 & & Role Consistency & Process Consistency & Contextual Coherence & Basic Text Quality \\
\hline
\multirow{5}{*}{Human} 
& CoRPS-Dial & 9.17 ± 0.38 & 9.43 ± 0.29 & 9.17 ± 0.38 & 9.56 ± 0.22 \\
& Raw Doctor & 8.47 ± 0.62 & 7.61 ± 0.83 & 7.94 ± 0.67 & 8.83 ± 0.51 \\
& Free Patient & 7.93 ± 0.71 & 9.11 ± 0.37 & 8.32 ± 0.59 & 9.12 ± 0.36 \\
& Single Agent & 8.50 ± 0.68 & 6.28 ± 1.76 & 7.70 ± 0.79 & 8.64 ± 0.56 \\
& Minimal Agents & 6.18 ± 1.24 & 5.84 ± 1.36 & 6.03 ± 1.41 & 7.41 ± 1.08 \\
& No QC/SR & 8.95 ± 0.49 & 9.03 ± 0.21 & 8.71 ± 0.56 & 8.92 ± 0.47 \\
\hline
\multirow{5}{*}{Deepseek-v2} 
& CoRPS-Dial & 9.08 ± 0.41 & 9.36 ± 0.33 & 9.02 ± 0.44 & 9.44 ± 0.28 \\
& Raw Doctor & 8.22 ± 0.73 & 7.28 ± 0.91 & 7.63 ± 0.74 & 8.61 ± 0.59 \\
& Free Patient & 7.66 ± 0.85 & 8.97 ± 0.44 & 8.06 ± 0.66 & 8.97 ± 0.42 \\
& Single Agent & 8.02 ± 0.14 & 8.84 ± 0.47 & 7.80 ± 0.57 & 8.72 ± 0.45 \\
& Minimal Agents & 6.44 ± 1.11 & 6.02 ± 1.28 & 6.29 ± 1.32 & 7.66 ± 0.96 \\
& No QC/SR & 8.73 ± 0.58 & 9.22 ± 0.39 & 8.48 ± 0.63 & 8.74 ± 0.55 \\
\hline
\end{tabular}
\caption{Human and Deepseek-v2 evaluation results compared with baselines across role consistency, process consistency, contextual coherence, and basic text quality. ``xxx ± xxx'' denotes mean ± std.}
\label{tab:dual-abl}
\end{table*}

\subsection{Datasets}
We collected 256 real-world pre-consultation dialogues from the Department of Cardiology, covering chief complaints, present illness history, and relevant risk factors for patients suspected of CHD. The pre-consultation phase focused on collecting information, without involving final diagnostic confirmation. We extracted these dialogues as seed patient profiles and augmented them via controlled perturbations along emotional state and communication style, generating 5,120 profiles. Each profile generated 1–3 multi-turn dialogues, resulting in a final 9,714 dialogues after post-hoc quality control.
Finally, we mix synthetic and real dialogues and split them into train/valid/test sets with an 8:1:1 ratio for downstream fine-tuning experiments.

\subsection{Evaluation Metrics}
We evaluate the alignment between synthetic and real pre-consultation dialogues from three perspectives: \textbf{(1) Statistical Distribution Consistency}, including Turns/Dialog, Sentence Length, Unique Chars, and Tokens/Turn; \textbf{(2) Semantic Consistency}, including Pairwise Dialogue Similarity, Paired Dialogue Similarity, and Local Coherence; and \textbf{(3) Lexical Diversity}, including Dist-1, Dist-2, Entropy, and Total Words (details in Appendix~\ref{metric-details}). In addition, we randomly sample 50 real and synthetic dialogues for human and LLM-based evaluation (rubric in Table~\ref{tab:glm-score}, Appendix~\ref{glm4-score}).

% To assess the practical utility of the synthetic data, 
We further evaluate downstream dialogue performance in terms of effective coverage, efficiency, turns, context utilization, repetition rate, invalid question rate, termination ability, and LLM-based evaluation (details in Appendix~\ref{downstream-metric-details}).

\begin{table*}
\centering
\scriptsize
\renewcommand{\arraystretch}{1.1}
\begin{tabular}{l c c c c c c c}
\hline
Metric & Real (256) & Syn (256) & Syn (1024) & Syn (2048) & Syn (4096) & Syn (8192) & Syn (9714) \\
\hline
\multicolumn{8}{c}{\textbf{Structure}} \\
\hline
Turns/Dialog & 34.26 ± 12.3 & 42.51 ± 4.32 & 42.39 ± 4.32 & 42.33 ± 4.46 & 42.33 ± 4.49 & 42.24 ± 4.48 & 42.25 ± 4.50 \\
Sent Len & 33.53 ± 74.29 & 57.60 ± 64.80 & 58.30 ± 65.81 & 56.88 ± 64.18 & 57.02 ± 64.51 & 57.22 ± 64.81 & 57.29 ± 64.95 \\
Unique Chars & 1413 & 1510 & 1837 & 1999 & 2162 & 2311 & 2350 \\
Tokens/Turn & 23.64 & 37.05 & 37.42 & 36.48 & 36.60 & 36.72 & 36.77 \\
\hline
\multicolumn{8}{c}{\textbf{Semantic}} \\
\hline
Pairwise Sim & - & 0.9240 & 0.9237 & 0.9245 & 0.9246 & 0.9244 & 0.9244 \\
Paired Sim & - & 0.9223 & - & - & - & - & - \\
Local Coh & 0.5363 ± 0.0234 & 0.5933 ± 0.0275 & 0.5946 ± 0.0268 & 0.5944 ± 0.0258 & 0.5945 ± 0.0257 & 0.5938 ± 0.0260 & 0.5938 ± 0.0261 \\
\hline
\multicolumn{8}{c}{\textbf{Diversity}} \\
\hline
Dist-1 & 0.0385 & 0.0426 & 0.0251 & 0.0192 & 0.0120 & 0.0092 & 0.0090\\
Dist-2 & 0.1910 & 0.2209 & 0.1210 & 0.1046 & 0.0635 & 0.0556 & 0.0542\\
Entropy & 8.2250 & 8.2467 & 8.2761 & 8.2665 & 8.2713 & 8.2639 & 8.2748\\
Total Words & 170,852 & 401,712 & 1,619,987 & 3,153,325 & 6,325,195 & 12,669,723 & 15,046,589 \\
\hline
\end{tabular}
\caption{Evaluation of dialogue quality across structure, semantics, and diversity for synthetic and real dialogues. ``Syn'' means synthetic. ``Pairwise Sim'' means pairwise dialogue similarity. ``Paired Sim'' means paired dialogue similarity.}
\label{tab:quality-comparison}
\end{table*}

\subsection{Implementation Details}
Both Doctor and Patient agents are instantiated using Qwen3-32B \citep{qwen3}, with a temperature of $0.2$. The same model is used for patient persona construction and clinical process modeling, including profile extraction and inquiry transition mining for building the guideline-grounded rule tree.

The post-hoc quality control module combines rule-based validation with GLM-4-9B reward scoring. Dialogues with average reward below $\theta=7.2$ are selectively rewritten based on dialogue context, patient profiles, and error feedback. We set $T=0.0$ for scoring and $T=0.1$ for rewriting. All models operate in no-thinking mode.

We fine-tune LLaMA-3.1-8B \citep{meta2024llama31}, DeepSeek-R1-Distill-LLaMA-8B, and DeepSeek-R1-Distill-Qwen-14B with LoRA on the mixed dataset, using learning rates of $3\times10^{-4}$ (8B) and $2\times10^{-4}$ (14B).

For evaluation, we use DeepSeek-v2 \citep{deepseek, Interactive-Agents} as the LLM judge ($T=0.1$) for synthetic dialogue evaluation, GLM-4-9B for downstream evaluation, bge-large-zh-v1.5 \citep{bge} for semantic consistency, and Qwen2-7B-Instruct \citep{qwen2} for Tokens/Turn measurement. All experiments are conducted on NVIDIA A100 GPUs.

\subsection{Baseline}
% To rigorously evaluate our approach, 
We compare against the following variants: 1) \textbf{Raw Doctor}: a Doctor Agent using only the raw schema without structured clinical process modeling; 2) \textbf{Free Patient}: a Patient Agent without persona conditioning; 3) \textbf{Single Agent}: a single agent with both structured clinical process and patient persona; 4) \textbf{Minimal Agents}: dual agents without structured process or persona (least constrained); 5) \textbf{No QC/SR}: removing rule-based and reward-based quality control and disabling selective rewriting.

For downstream evaluation, we further compare model performance before and after supervised fine-tuning (SFT) to assess the impact of synthetic data, including LLaMA-3.1-8B vs.\ LLaMA-3.1-8B-SFT, DeepSeek-R1-Distill-LLaMA-8B vs.\ its SFT variant, and DeepSeek-R1-Distill-Qwen-14B vs.\ its SFT variant.

\section{Experimental Results}
\label{exper:res}

\subsection{Main Results}
To evaluate the effectiveness of our framework, we compare CoRPS-Dial with multiple baselines across structural, semantic, and diversity metrics (Table~\ref{tab:abl}), and further assess dialogue quality using human and LLM-based evaluation (Table~\ref{tab:dual-abl}).
Overall, CoRPS-Dial consistently outperforms all baselines, generating more structured, coherent, and diverse dialogues. Raw Doctor produces shorter dialogues with degraded semantic coherence, highlighting the importance of explicit clinical workflow modeling. Free Patient shows reduced lexical diversity and contextual richness, indicating the need for persona conditioning to achieve realistic patient variability. Single Agent generates fluent but less diverse dialogues with more monotonous interaction patterns, while Minimal Agents exhibit unstable role transitions and poor coherence despite higher surface-level variation. No QC/SR introduces more verbosity and textual inconsistencies, demonstrating the importance of quality control and selective refinement.

By integrating structured clinical workflows, personalized patient personas, dual-agent interaction, and quality control with selective refinement, CoRPS-Dial achieves the best performance in both human and DeepSeek-v2–based evaluation, demonstrating superior role consistency, process fidelity, contextual coherence, and linguistic quality.

\begin{table*}
\centering
\midfont
\begin{tabular}{
l
>{\centering\arraybackslash}m{1.5cm}
>{\centering\arraybackslash}m{1.5cm}
>{\centering\arraybackslash}m{1.5cm}
>{\centering\arraybackslash}m{2cm}
>{\centering\arraybackslash}m{1.5cm}
>{\centering\arraybackslash}m{2cm}
}
\hline
Metric & LLaMA-8B & LLaMA-8B-SFT & DeepSeek-R1-8B & DeepSeek-R1-8B-SFT & DeepSeek-R1-14B & DeepSeek-R1-14B-SFT \\
\hline
Effective Coverage $\uparrow$ & 16.48\% & 46.83\% & 36.2\% & 23.17\% & 29.33\% & 44.65\% \\
Efficiency $\uparrow$ & 0.75 & 2.13 & 1.65 & 1.05 & 1.33 & 2.03 \\
Turns & 22 & 21.98 & 22 & 22 & 22 & 22 \\
Context Utilization $\uparrow$ & 23.35\% & 26.28\% & 27.13\% & 38.25\% & 26.86\% & 23.73\% \\
Repetition Rate $\downarrow$ & 79.64\% & 19.33\% & 27.52\% & 66.60\% & 30.57\% & 7.55\% \\
Invalid Question Rate $\downarrow$ & 90.12\% & 38.67\% & 38.52\% & 75.69\% & 45.27\% & 32.36\% \\
Termination Ability & 0\% & 51\% & 0\% & 0\% & 0\% & 0.5\% \\
\hline
\multicolumn{7}{c}{\textbf{LLM-based Evaluation}} \\
\hline
Coherence & 7.18 & 8.22 & 8.45 & 7.75 & 8.25 & 8.20 \\
Relevance & 6.31 & 7.29 & 7.52 & 6.75 & 7.38 & 7.29 \\
Clinical Logic & 7.94 & 8.83 & 8.83 & 8.13 & 8.77 & 8.81 \\
Informativeness & 5.51 & 6.29 & 6.52 & 5.80 & 6.39 & 6.32 \\
Fluency & 8.15 & 8.30 & 8.44 & 8.41 & 8.36 & 8.28 \\
\hline
\end{tabular}
\caption{Performance comparison of three models before and after SFT. $\uparrow$ means higher values are better, and $\downarrow$ means lower values are better.}
\label{tab:model_performance}
\end{table*}

\begin{figure}[t]
    \centering
    \includegraphics[width=0.90\columnwidth]{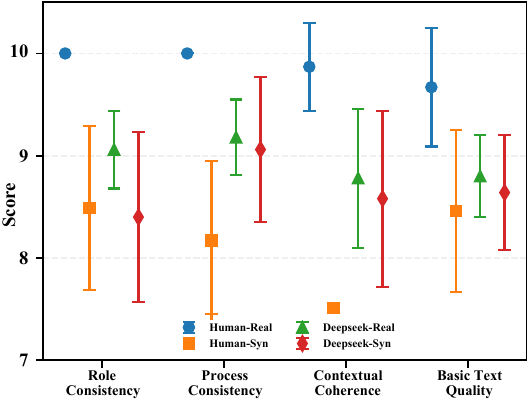}
    \caption{Human and LLM-as-a-Judge evaluations of 50 real and 50 synthetic doctor–patient dialogues on a 10-point Likert scale. Human scores are averaged over 3 domain experts (mean ± std), and LLM scores are from Deepseek-v2 across 3 epochs (mean ± std).}
    \label{fig:Human-llm}
\end{figure}

\subsection{Quality and Scalability of Synthetic Dialogues} 
We evaluate the quality and scalability of synthetic dialogues across different sizes by sampling subsets of 256, 1024, 2048, 4096, and 8192 from the full synthetic corpus and compare them with real-world clinical dialogues (Table~\ref{tab:quality-comparison}).
Detailed analyses are provided in Appendix~\ref{quality-analysis}.

Overall, synthetic dialogues maintain stable structural and semantic quality across different sizes, demonstrating strong robustness and scalability. Although increasing scale introduces mild surface-level repetition, lexical richness and semantic coherence are well preserved, validating the suitability of large-scale synthetic dialogues for downstream model training.

Moreover, to assess the impact of emotion and communication styles, we select exactly one representative dialogue per portrait of the augmented patient profile, for a total of 5,120 dialogues. Results (Appendix~\ref{quality-analysis}) confirm that statistical distribution consistency, semantic consistency, and lexical diversity are preserved across variants, and no systematic repetition or distortion appears in the chief complaints, indicating that such controlled stylistic augmentation effectively enriches the dialogue while maintaining high clinical reliability.

\subsection{Human and LLM Evaluation of Real and Synthetic Dialogue Quality}
Figure~\ref{fig:Human-llm} summarizes evaluations on 50 real and 50 synthetic doctor–patient dialogues. DeepSeek-v2 assigns uniformly high scores to both real (8.78–9.18) and synthetic data (8.40–9.06), showing limited discrimination (<0.4 variance). In contrast, human experts produce a clear gap: real dialogues reach near-ceiling scores (9.67–10), while synthetic ones drop to 7.51–8.49, with differences up to 2.0 points.
We further report Cohen’s kappa for human--LLM agreement in Appendix~\ref{human-llm-agreement}, observing strong consistency across all dimensions. These results suggest that while synthetic dialogues are structurally robust, LLM-based evaluation underestimates subtle quality deficiencies that human judgment captures.

\subsection{Downstream Validation}
To assess the practical utility of synthetic data, we present the performance of models before and after SFT in Table~\ref{tab:model_performance}. Our results demonstrate that the synthetic data leads to substantial improvements in downstream tasks across most models, with consistent gains observed in coverage, efficiency, and dialogue control. Notably, we observe significant enhancements in models such as LLaMA-8B and DeepSeek-R1-14B. These improvements suggest that synthetic dialogues provide valuable supervision, particularly when aligned with model capacity and appropriate training configurations. Our findings reinforce that synthetic data is both effective and transferable, though the benefits are maximized with tailored fine-tuning strategies specific to each model architecture.

\subsection{Case Study}
We qualitatively compare dialogues generated by all baselines in Appendix~\ref{case-study}. Existing methods exhibit issues such as insufficient symptom inquiry, limited persona diversity, unstable interaction patterns, and textual inconsistencies. In contrast, CoRPS-Dial produces more consistent roles, coherent interaction flows, and cleaner language through a structured generation process and post-hoc quality refinement.

\section{Conclusion}
We present CoRPS-Dial, a framework for generating high-quality clinical dialogues by combining structured workflow, personalized patient personas, dual-agent interaction, and two-stage post-hoc quality control. Our method addresses role inconsistency, process redundancy, semantic drift, and textual noise in synthetic dialogue generation. Evaluations show CoRPS-Dial outperforms all baselines, producing coherent, clinically faithful, and diverse dialogues. Case studies further demonstrate improved role consistency and workflow fidelity. CoRPS-Dial offers a scalable, reliable approach to constructing clinical dialogues, enabling future research on controllable, data-centric medical conversational systems.

\section*{Limitations}
While CoRPS-Dial shows strong performance in generating clinically grounded dialogues, there remains room for further improvement and extension. First, hallucination in LLMs remains difficult to eliminate; future work may improve reliability by incorporating expert validation or external knowledge. Second, current validation focuses on CHD, and extending it to other diseases could assess generalizability. Third, using single-institution data may introduce bias; incorporating multi-center data could enhance diversity. Lastly, although the framework is aimed at academic research, it has potential for future clinical deployment.

\section*{Ethical Considerations}
This study involves real-world clinical pre-consultation dialogues and was conducted in accordance with institutional regulations for human-subject research and medical data governance. All participants provided informed consent prior to recording. To protect privacy, all data were de-identified before processing by removing or replacing personally identifiable information and assigning unique coded identifiers. Audio was recorded using institution-approved devices and transcribed via a locally deployed speech-to-text system to prevent external transmission. Transcripts and metadata were stored on encrypted institutional servers with strict access control, and only authorized researchers could access the raw data. Synthetic dialogues were generated exclusively from de-identified seed profiles using controlled perturbations of clinical attributes, emotional states, and communication styles. The generation process preserves statistical properties of clinical interactions without reconstructing identifiable patient information, thereby minimizing re-identification risk. The resulting dialogues and models are intended solely for research and methodological evaluation; due to potential inaccuracies or hallucinations in generative models, they are not designed for direct clinical decision-making without professional oversight.

% \section*{Acknowledgments}
% This work was supported by the the National Natural Science Foundation of China under Grant No. 62576254 and the National Key Research and Development Project of China under Grant No. 2024YFC3308400.

% Finally, we want to thank Beier Nan, Yifan Kuai, Yuchen Xiong, Kangkang Tian, Jia Li, Tian Bai for their help with data collection and data filtering, as well as Yuchen Xiong, Kangkang Tian, Xu Zhu, Jia Li for their contributions to data quality checking.


\bibliography{custom}

\appendix

\section{Appendix}
\label{sec:appendix}
\subsection{Noise Handling and Structural Normalization of Real-World Clinical Dialogues}
\label{nosie}
This section focuses on procedures for surface-level noise filtering arising from speech disfluencies, interruptions, and non-clinical content. All filtering is applied conservatively to ensure that clinically relevant information remains. The filtering steps includes:

\textbf{Step 1: Disfluency removal.}
\begin{itemize}
    \item Eliminate filler words (e.g., “um”, “uh”) and self-corrections that do not affect clinical meaning.
    \item Preserve repeated mentions of symptoms if they convey emphasis.
\end{itemize}

\textbf{Step 2: Non-informative turn removal.}
\begin{itemize}
    \item Discard speaker turns that consist solely of social interactions, greetings, or unrelated chit-chat.
    \item Short utterances (< 3 tokens) without medical entities are considered for removal.
\end{itemize}

\textbf{Step 3: Preservation principle.}
\begin{itemize}
    \item Clinically relevant tokens, symptom mentions, and temporal indicators are retained.
    \item The filtering process does not modify the semantic content of the patient’s report.
\end{itemize}

\begin{figure}[t]
  \includegraphics[width=\columnwidth]{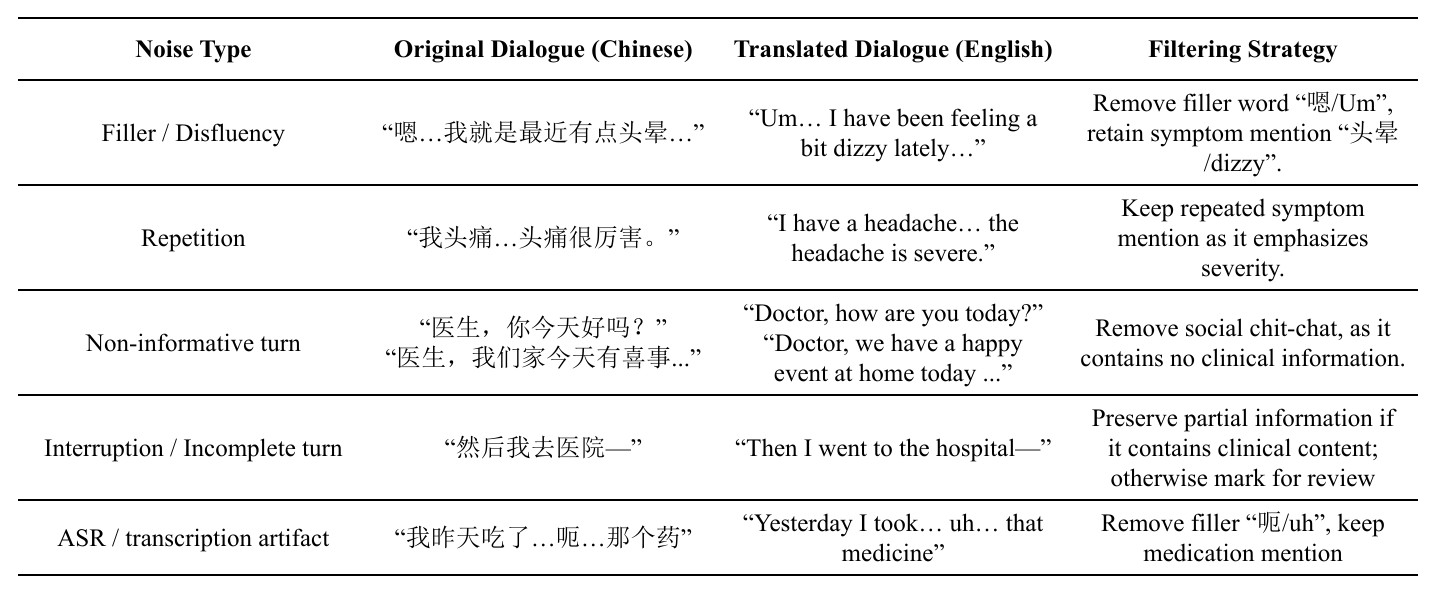}
  \caption{Examples of surface-level noise filtering in clinical dialogues.}
  \label{fig:noise}
\end{figure}

Examples are provided in Figure~\ref{fig:noise}.
% Examples are provided in the \textit{GitHub} repository.

% \begin{figure*}
%     \centering
%     \includegraphics[width=0.8\textwidth]{Figures/noise.pdf}
%     \caption{Examples of surface-level noise filtering in clinical dialogues.}
%     \label{fig:noise}
% \end{figure*}

\subsection{Scoring Criteria for Reward Evaluation}
\label{glm4-score}
This section presents the scoring criteria used by GLM-4-9B to evaluate synthetic dialogues (Table \ref{tab:glm-score}).

\begin{table*}
\centering
\scriptsize
% \begin{tabular}{c>{\raggedright\arraybackslash}p{7cm}c>{\raggedright\arraybackslash}p{7cm}}
\renewcommand{\arraystretch}{1.1}
\begin{tabular}{c>{\raggedright\arraybackslash}p{0.35\textwidth}c>{\raggedright\arraybackslash}p{0.35\textwidth}}

% \begin{tabular}{cccc}
\hline
Type & Scoring criteria & Score & Scoring Rule \\
\hline
\multirow{17}{*}{Role Consistency} 
& Perfect consistency, where each role behaves completely according to its expected identity and duties, displaying professional and appropriate actions throughout. & 10 & Flawless role adherence.\\
& High consistency, where most dialogues align with role expectations, with only minor deviations in a few & 7-9 & 9: No role violation, only trivial phrasing imperfections may occur. \\
& instances. && 8: 1--2 minor out-of-character instances (e.g., patient repeats medical terms). \\
&&& 7: $\ge$3 minor or 1 moderate deviation (e.g., patient self-diagnosis).\\
& Minor inconsistencies, where most dialogues are role-appropriate, but occasional deviations occur. & 4-6 & 6: Clear role overstepping but primary roles remain identifiable.\\
& && 5: Frequent role overstepping ($>$3 moderate instances).\\
&&& 4: Severe role confusion (e.g., patient leads questioning).\\
& Significant inconsistencies, where the roles are clearly violated. & 1-3 & 3: Partial role reversal (patient fully acts as doctor, or vice versa). \\
&&& 2: Complete role reversal (content irrelevant to roles).\\
&&& 1: Role collapse (identities indistinguishable).\\
\hline
\multirow{15}{*}{Process Consistency} 
& Perfect process alignment, where all steps strictly follow the required order in $\mathcal{T}$, and all red flags and dependencies are properly respected. & 10 & No omission/repetition/order error. \\
& Mostly consistent process, with only minor deviations in the sequence. & 7-9 & 9: 1--2 minor issues (e.g., non-mandatory item slightly out of order, or harmless repetition). \\
&&& 8: 3--4 minor / 1 repetition / 1 non-mandatory omitted. \\
&&& 7: $\ge$5 minor / 1 mandatory indirectly omitted / delayed red flag. \\
& Some process inconsistencies, where most steps follow the correct order, but a few key nodes or red flags are & 4-6 & 6: 1 mandatory omitted / 1 order error / 2 red flags missed.\\
& overlooked or the order of some steps is incorrect. && 5: 2 mandatory omitted / 3 red flags / $\ge$3 repetitions. \\
& && 4: 3 mandatory omitted / phase skipped. \\
& Severe process violations, where the dialogue is & 1-3 & 3: 4--5 mandatory omitted / severe fragmentation. \\
& completely out of order, multiple key nodes or red flags && 2: 6+ mandatory omitted / key phase missing. \\
& are missed, and dependencies are ignored. && 1: No recognizable structure. \\
\hline
\multirow{22}{*}{Contextual Coherence}
& Perfect coherence, with no contradictions in symptoms, timelines, or medical logic. The dialogue shows complete internal consistency. & 10 & No contradictions. \\
& Overall coherence, with only minor, isolated contradictions or unclear medical logic that do not affect the overall consistency. & 7-9 & 9: 1 minor contradiction (e.g., symptom severity slightly inconsistent across turns) that does not affect understanding.\\
&&& 8: 2 minor / 1 unclear logic (e.g., vague temporal relationship).\\
&&& 7: 3 minor / 1 moderate contradiction (e.g., symptom onset time shifts by >1 month). \\
& Occasional minor contradictions, where some symptoms & 4-6 & 6: 1 moderate+1 minor / 2 moderate / 1 clear logic error.\\
& or timelines may slightly differ, or there are small logical inconsistencies in medical facts. && 5: 3 moderate / 1 major contradiction (e.g., symptom location changes completely between turns without explanation). \\
&&& 4: 2 major contradictions, or a combination of contradictions that significantly undermine symptom/timeline credibility. \\
& Significant contradictions or inconsistencies, where symptom descriptions or timelines do not match, or medical logic is completely flawed. & 1-3 & 3: 3 major contradictions, or entire symptom description contradicts itself (e.g., "never had chest pain" vs. "chest pain for 2 years"). \\
&&& 2: Multiple major contradictions across dimensions. \\
&&& 1: Complete inconsistency.\\
\hline
\multirow{19}{*}{Basic Text Quality}
& Perfect text quality, with no formatting errors, incorrect medical terminology, or violations of safety guidelines. & 10 & No errors.\\
& High-quality text with occasional minor formatting or & 7-9 & 9: 1 minor error (e.g., typo). \\
& medical term errors, but the content largely adheres to guidelines. && 8: 2--3 minor / 1 minor terminology (e.g., "stomach ache"). \\
&&& 7: 4--5 minor / 1 moderate terminology (e.g., "heart attack") / 1 minor safety (e.g., vague dosing). \\
& Moderate text quality, with some formatting issues, incorrect medical terms, or small guideline violations, but the content is still understandable. & 4-6 & 6: 6--8 minor / 2 moderate terminology / 1 major terminology (e.g., "hypertension" vs "hyperthyroidism") / 1 moderate safety (e.g., over-the-counter without checking). \\
&&& 5: 9--12 minor / 3 moderate / 2 major terminology / 1 severe safety (e.g., contraindicated med). \\
&&& 4: 13+ minor / multiple moderate safety violations. \\
& Severe text quality issues, with frequent errors, unsafe content, or major guideline violations. & 1-3 & 3: Critical safety violations (e.g., stop essential med) but partially understandable. \\
&&& 2: Barely understandable / multiple critical safety violations. \\
&&& 1: Gibberish / actively dangerous content. \\
\hline
\end{tabular}
\caption{Specific scoring criteria for GLM-4-9B. We assess the quality of the synthetic dialogues in terms of role consistency, process consistency, contextual coherence, and basic text quality, then score them according to different criteria, and finally average them.}
\label{tab:glm-score}
\end{table*}

\subsection{Details of Evaluation Metrics}
\label{metric-details}
This section presents the details of evaluation metrics, as follows:

\textbf{(1) Statistical Distribution Consistency.} 

\begin{itemize}
    \item Turns / Dialog: average number of speaker turns per consultation.
    \item Sentence Length: average length of sentence (in characters) per consultation.
    \item Unique Chars: vocabulary size after removing common punctuation and whitespace.
    \item Tokens / Turn: average tokenizer word pieces per turn.
\end{itemize}

\textbf{(2) Semantic Consistency.} 

\begin{itemize}
    \item Pairwise Similarity: mean cosine similarity between the real and synthetic.
    \item Paired Similarity: mean cosine similarity between turns in a dialogue.
    \item Local Coherence: average cosine similarity between embeddings of consecutive utterances within each dialogue.
\end{itemize}

\textbf{(3) Lexical Diversity.}

\begin{itemize}
    \item Dist-1: number of unique unigrams (single words) divided by the total number of unigrams in the generated text.
    \item Dist-2: number of unique bigrams (pairs of consecutive words) divided by the total number of bigrams in the generated text.
    \item Entropy: average word-level entropy of the generated text, where higher values indicate more uniform word distribution and greater lexical richness.
    \item Total Words: total number of words (tokens) produced across all generated text.
\end{itemize}

\subsection{Downstream Evaluation Metrics}
\label{downstream-metric-details}
This section defines the metrics for evaluating downstream dialogue performance. The following metrics are considered:

\begin{itemize}
    \item \textbf{Effective Coverage} measures whether the model captures key medical information.
    \item \textbf{Efficiency} measures the model’s ability to gather sufficient information within limited dialogue turns.
    \item \textbf{Context Utilization} measures whether the model leverages the patient’s most recent responses to guide follow-up questions.
    \item \textbf{Repetition Rate} measures the extent of repeated or redundant questioning.
    \item \textbf{Invalid Question Rate} measures the proportion of irrelevant or unnecessary questions generated by the model.
    \item \textbf{Termination Ability} measures whether the model appropriately concludes the consultation once sufficient information is collected.
    \item \textbf{LLM-based Evaluation} measures overall consultation quality using an independent LLM, including \textit{Coherence} (logical consistency), \textit{Relevance} (alignment with clinical context), \textit{Clinical Logic} (medical correctness and reasoning), \textit{Informativeness} (usefulness of information), and \textit{Fluency} (naturalness and readability).
\end{itemize}

\subsection{Quality and Scalability Analysis of Synthetic Dialogues}
\label{quality-analysis}
\paragraph{Quality and Scalability Across Dataset Sizes.}

We provide a detailed analysis of the structural, semantic, and diversity properties of synthetic dialogues across different dataset sizes, as shown in Table~\ref{tab:quality-comparison}.

\textbf{Structural Properties.} Real-world dialogues exhibit substantial variability in turns per dialogue (34.26 ± 12.3) and sentence length (33.53 ± 74.29), reflecting the flexible nature of human clinical conversations. In contrast, synthetic dialogues show more stable structural patterns across all scales, with approximately 42 turns per dialogue and significantly lower variance. Synthetic dialogues also contain longer sentences and more tokens per turn, indicating more explicit and detailed utterances. The number of unique characters increases steadily with size, suggesting growing lexical coverage.

\textbf{Semantic Consistency.} Synthetic dialogues achieve consistently high semantic similarity across all scales (Pairwise Sim $\approx$ 0.924, Paired Sim $\approx$ 0.922), demonstrating that semantic structure is preserved even in small subsets. Local coherence is also higher than in real-world dialogues (0.593–0.594 vs. 0.536), indicating that structured generation enhances turn-level coherence. Notably, these semantic metrics remain stable as size increases, suggesting that quality does not degrade with scale.

\textbf{Diversity Analysis.} While real-world dialogues show higher Dist-1 and Dist-2 due to unconstrained language use, synthetic dialogues exhibit a gradual decrease in these metrics as size increases, reflecting repeated clinical patterns. However, lexical entropy remains stable (8.26–8.27), indicating preserved lexical richness. Total word count grows linearly with size, confirming that large-scale synthetic generation expands textual resources for downstream training.

% \subsection{Quality Analysis of Dialogues Generated from Augmented Patient Profiles}
% \label{qua-aug}
\paragraph{Analysis of Augmented Patient Profiles.}

To assess the impact of emotion and communication styles, we select exactly one representative dialogue per portrait of the augmented patient profile, for a total of 5,120 dialogues, as shown in Table \ref{tab:quality-aug}.
% Detailed experimental results are available in the \textit{GitHub} repository.
% , as shown in Table \ref{tab:quality-aug}.

\begin{table*}
\centering
\midfont
\renewcommand{\arraystretch}{1.1}
\begin{tabular}{l c c c c c c c}
\hline
Metric & Real (256) & Syn (256) & Syn (1024) & Syn (2048) & Syn (4096) & Syn (5120) \\
\hline
\multicolumn{7}{c}{\textbf{Structure}} \\
\hline
Turns/Dialog & 34.26 ± 12.30 & 42.18 ± 4.26 & 42.50 ± 4.40 & 42.10 ± 4.53 & 42.16 ± 4.52 & 42.14 ± 4.52 \\
Sent Len & 33.53 ± 74.29 & 56.41 ± 63.78 & 57.82 ± 64.68 & 58.15 ± 65.33 & 57.74 ± 65.33 & 57.66 ± 65.20 \\
Unique Chars & 1413 & 1514 & 1861 & 2014 & 2164 & 2211 \\
Tokens/Turn & 23.64 & 36.13 & 37.11 & 37.27 & 37.03 & 36.98 \\
\hline
\multicolumn{7}{c}{\textbf{Semantic}} \\
\hline
Pairwise Sim & - & 0.9256 & 0.9242 & 0.9239 & 0.9241 & 0.9241 \\
Paired Sim & - & 0.9255 & - & - & - & - \\
Local Coh & 0.5363 ± 0.0234 & 0.5930 ± 0.0260 & 0.5960 ± 0.0256 & 0.5953 ± 0.0260 & 0.5945 ± 0.0260 & 0.5943 ± 0.0260 \\
\hline
\multicolumn{7}{c}{\textbf{Diversity}} \\
\hline
Dist-1 & 0.0385 & 0.0429 & 0.0250 & 0.0121 & 0.0090 & 0.0088 \\
Dist-2 & 0.1910 & 0.2201 & 0.1211 & 0.1048 & 0.0637 & 0.0558 \\
Entropy & 8.2250 & 8.2537 & 8.2664 & 8.2739 & 8.2735 & 8.2751 \\
Total Words & 170,852 & 389,320 & 1,610,047 & 3,206,744 & 6,379,972 & 7,989,374 \\
\hline
\end{tabular}
\caption{Comparison of augmented profile-based dialogue generation quality in terms of structure, semantics, and diversity, with random sampling from augmented patient profiles, selecting one dialogue per profile. ``Syn'' means synthetic, ``Sent Len'' means sentence length. ``Pairwise Sim'' means pairwise dialogue similarity. ``Paired Sim'' means paired dialogue similarity. ``Local Coh'' means local coherence.}
\label{tab:quality-aug}
\end{table*}

\textbf{Structural and Semantic Consistency.} Synthetic dialogues maintain consistent turns per dialogue (42) and sentence length across all sampling scales, with low variance compared to real-world dialogue. Local coherence and pairwise similarity remain high (0.593–0.596), indicating that varying emotion and expression style do not disrupt the underlying semantic structure. These results demonstrate that dialogue expansion through style variation preserves the integrity of the clinical content while generating diverse surface realizations.

\textbf{Diversity and Robustness.} While n-gram diversity (Dist-1/Dist-2) decreases slightly as the number of dialogues per profile increases, overall lexical entropy remains stable (8.25–8.28). This shows that the variations in emotion and expression style introduce linguistic diversity without creating bias or artificial repetition in the core clinical information. The linear growth in total words further confirms that the expanded scales are appropriately expanded.

These findings suggest that profile-based expansion is a safe and effective strategy to increase size, introducing natural variability in patient expression while preserving clinical fidelity. Importantly, small subsets of the expanded dialogues reliably reflect the quality of the full dialogues, supporting their use in downstream dialogue modeling without introducing bias from repeated core information.

\subsection{Human--LLM Agreement Analysis}
\label{human-llm-agreement}
\begin{figure}[t]
    \centering
    \includegraphics[width=\columnwidth]{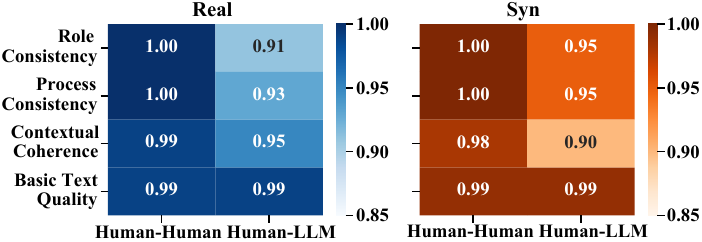}
    \caption{Cohen’s kappa inter-rater agreement across four evaluation dimensions for real and synthetic doctor–patient dialogues, comparing humans and human–LLM consistency.}
    \label{fig:kappa_heatmap}
\end{figure}

This section reports Cohen’s kappa scores for human--LLM agreement in Figure~\ref{fig:kappa_heatmap}, showing strong consistency across evaluation dimensions.

\subsection{Case Study}
\label{case-study}
\begin{figure}[t]
    \centering
    \begin{minipage}{\columnwidth}
        \centering
        \includegraphics[width=0.9\columnwidth]{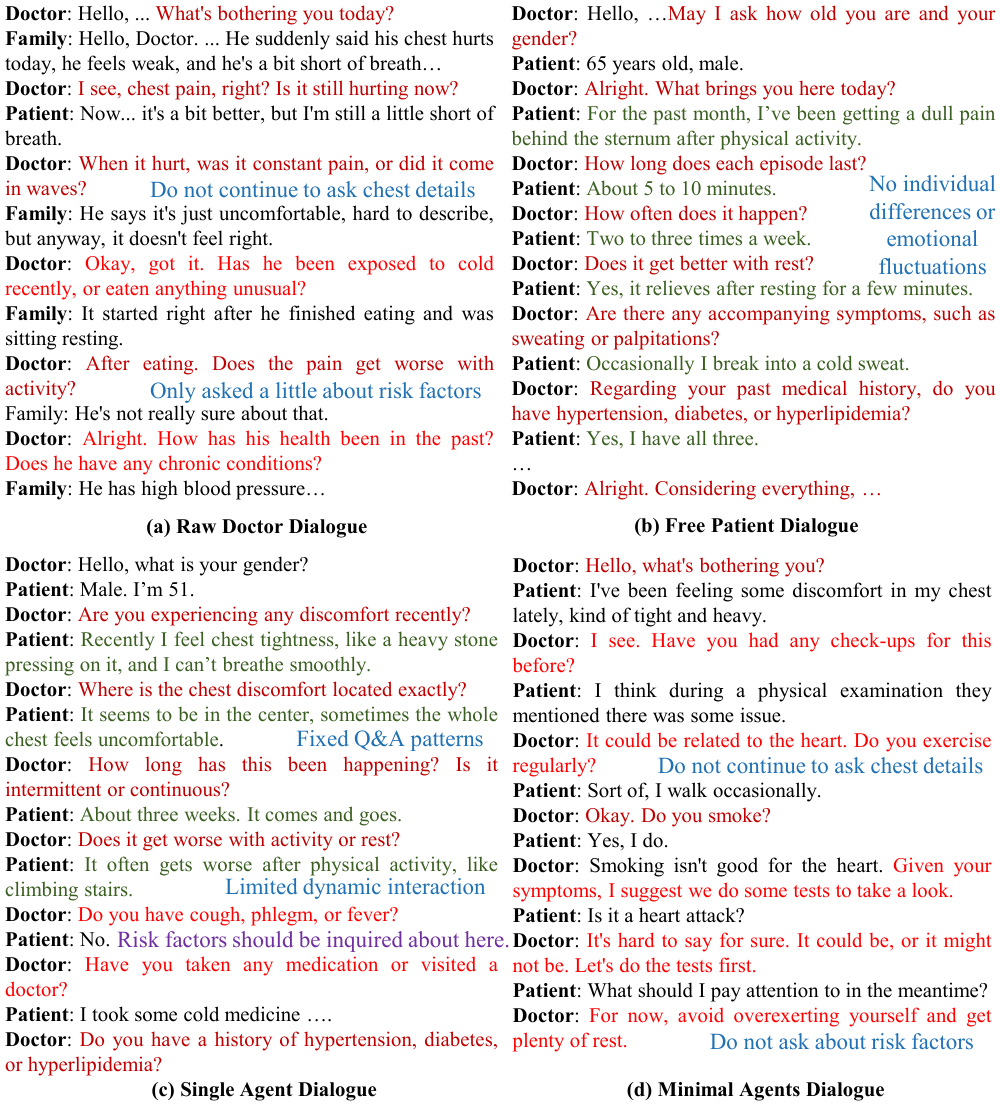}
        \vspace{0.05cm}
    \end{minipage}
    
    \begin{minipage}{\columnwidth}
        \centering
        \includegraphics[width=0.9\columnwidth]{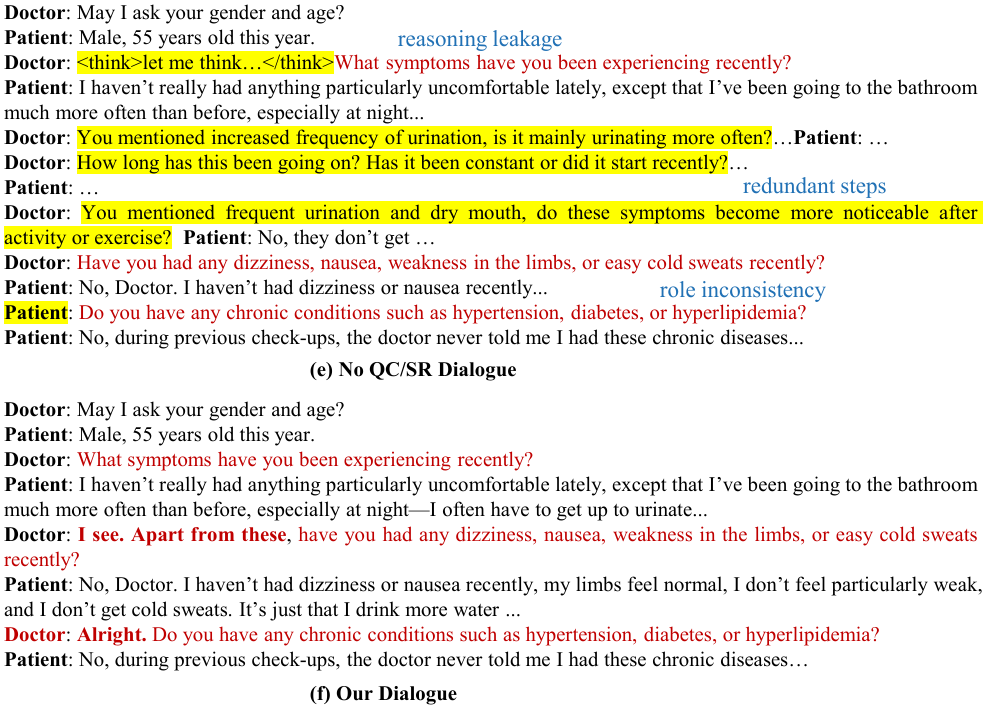}
    \end{minipage}
    
    \caption{Qualitative comparison of dialogues generated by baselines. Deep red indicates relevance, light red indicates weak relevance, and a yellow background indicates an issue.}
    \label{fig:case_study}
\end{figure}

Figure~\ref{fig:case_study} presents representative dialogues generated by different baselines and CoRPS-Dial.

\textbf{Raw Doctor} lacks targeted symptom and risk-factor inquiry. \textbf{Free Patient} produces template-like responses with limited persona diversity. \textbf{Single Agent} generates coherent but less diverse interactions, while \textbf{Minimal Agents} suffer from unfocused dialogues due to missing role and process constraints.

Although \textbf{No QC/SR} benefits from structured workflows, personas, and dual-agent interaction, it still exhibits role inconsistency, incomplete or redundant steps, and textual noise. Our method generates more consistent roles, coherent interaction flows, and cleaner language through post-hoc quality control and selective refinement.

\end{document}